\documentclass[runningheads]{llncs}
\usepackage{graphicx}
\usepackage{comment}
\usepackage{amsmath,amssymb} 
\usepackage{color}
\usepackage{multirow}
\usepackage{float}
\usepackage{booktabs}
\usepackage{hyperref}

\usepackage{microtype}

\newcommand{\Natural}{Internet}
\newcommand{\Symbolic}{Symbolic}
\newcommand{\Scenes}{Scenes \& Textures}
\newcommand{\Biological}{Biological}
\newcommand{\ra}[1]{\renewcommand{\arraystretch}{#1}}
\definecolor{mygray}{gray}{0.4}

\begin{document}
\pagestyle{headings}
\mainmatter
\def\ECCVSubNumber{5568}  

\title{Extending and Analyzing Self-Supervised Learning Across Domains} 

\newenvironment{packed_enum}{
\begin{enumerate}
  \setlength{\itemsep}{1pt}
  \setlength{\parskip}{0pt}
  \setlength{\parsep}{0pt}
}{\end{enumerate}}

\titlerunning{Self-Supervised Learning Across Domains}
%
\author{Bram Wallace \and
Bharath Hariharan }
\authorrunning{B. Wallace and B. Hariharan}
%
\institute{Cornell University \\ \email{bw462@cornell.edu bharathh@cs.cornell.edu} } 

\maketitle

\begin{abstract}
Self-supervised representation learning has achieved impressive results in recent years, with experiments primarily coming on ImageNet or other similarly large internet imagery datasets.
There has been little to no work with these methods on other smaller domains, such as satellite, textural, or biological imagery.
We experiment with several popular methods on an unprecedented variety of domains. 
We discover, among other findings, that Rotation is by far the most semantically meaningful task, while much of the performance of Jigsaw is attributable to the nature of its induced distribution rather than semantic understanding.
Additionally, there are several areas, such as fine-grain classification, where all tasks underperform. 
We quantitatively and qualitatively diagnose the reasons for these failures and successes via novel experiments studying pretext generalization, random labelings, and implicit dimensionality.
Code and models are available at \url{https://github.com/BramSW/Extending_SSRL_Across_Domains/}.
\end{abstract}

\section{Introduction}

A good visual representation is key to all visual recognition tasks.
However, in current practice, one needs large labeled training sets to train such a representation.
Unfortunately, such datasets can be hard to acquire in many domains, such as satellite imagery or the medical domain.
This is often either because annotations require expertise and experts have limited time, or the images themselves are limited (as in medicine).
To bring the benefits of visual recognition to these disparate domains, we need powerful representation learning techniques that do not require large labeled datasets.

A promising direction is to use self-supervised representation learning (SSRL), which has gained increasing interest over the last few years\cite{rotation,jigsaw,colorization,fair_benchmark,google_survey,split_ae}.
However, past work has primarily evaluated these techniques on general category object recognition in internet imagery (e.g. ImageNet classification)\cite{imagenet}.
There has been very little attention on how (and if) these techniques extend to other domains, be they fine-grained classification problems or datasets in biology and medicine.
Paradoxically, these domains are often most in need of such techniques precisely because of the lack of labeled training data.

As such, a key question is whether conclusions from benchmarks on self-supervised learning~\cite{fair_benchmark,google_survey} which focused on internet imagery, carry over to this broader universe of recognition problems.
In particular, does one technique dominate, or are different pretext tasks useful for different types of domains (\textbf{Sec.~\ref{sec:rankings}})?
Are representations from an ImageNet classifier still the best we can do (\textbf{Sec.~\ref{sec:rankings}})?
Do these answers change when labels are limited (\textbf{Sec.~\ref{sec:semi}})?
Are there problem domains where all proposed techniques currently fail (\textbf{Sec.~\ref{sec:failure}})?

A barrier to answering these questions is our limited understanding of self-supervised techniques themselves.
We have seen their \emph{empirical} success on ImageNet, but when they do succeed, what is it that drives their success (\textbf{Sec.~\ref{sec:success_method}})?
Furthermore, what does the space of learned representations look like, for instance in terms of the dimensionality (\textbf{Sec.~\ref{sec:pca}}) or nearest neighbors (\textbf{Sec.~\ref{sec:neighbors}})?


In this work, we take the first steps towards answering these questions.
We evaluate and analyze multiple self-supervised learning techniques (Rotation\cite{rotation}, Instance Discrimination\cite{inst_disc} and Jigsaw\cite{jigsaw}) on the broadest benchmark yet of 16 domains spanning internet, biological, satellite, and symbolic imagery.
We find that Rotation has the best overall accuracy (reflective of rankings on ImageNet), but is outperformed by Instance Discrimination on biological domains (\textbf{Sec.~\ref{sec:rankings}}).
When labels are scarce, pretext methods outperform ImageNet initialization and even full supervision on numerous tasks (\textbf{Sec.~\ref{sec:semi}}).
A prominent failure case for SSRL is fine-grained classification problems, due to important cues such as color being discarded during training (\textbf{Sec.~\ref{sec:failure}}).
Finally, when SSRL techniques do succeed, their reason for success varies: Rotation relies more on the semantic nature of the pretext task, compared to Jigsaw and Instance Discrimination (\textbf{Sec.~\ref{sec:success_method}}). Perhaps as a consequence, the representations of Rotation having comparatively higher implicit dimensionality (\textbf{Sec.~\ref{sec:pca}}).


\section{Datasets}

\begin{figure}[!tbh]
\centering
\includegraphics[width=\linewidth]{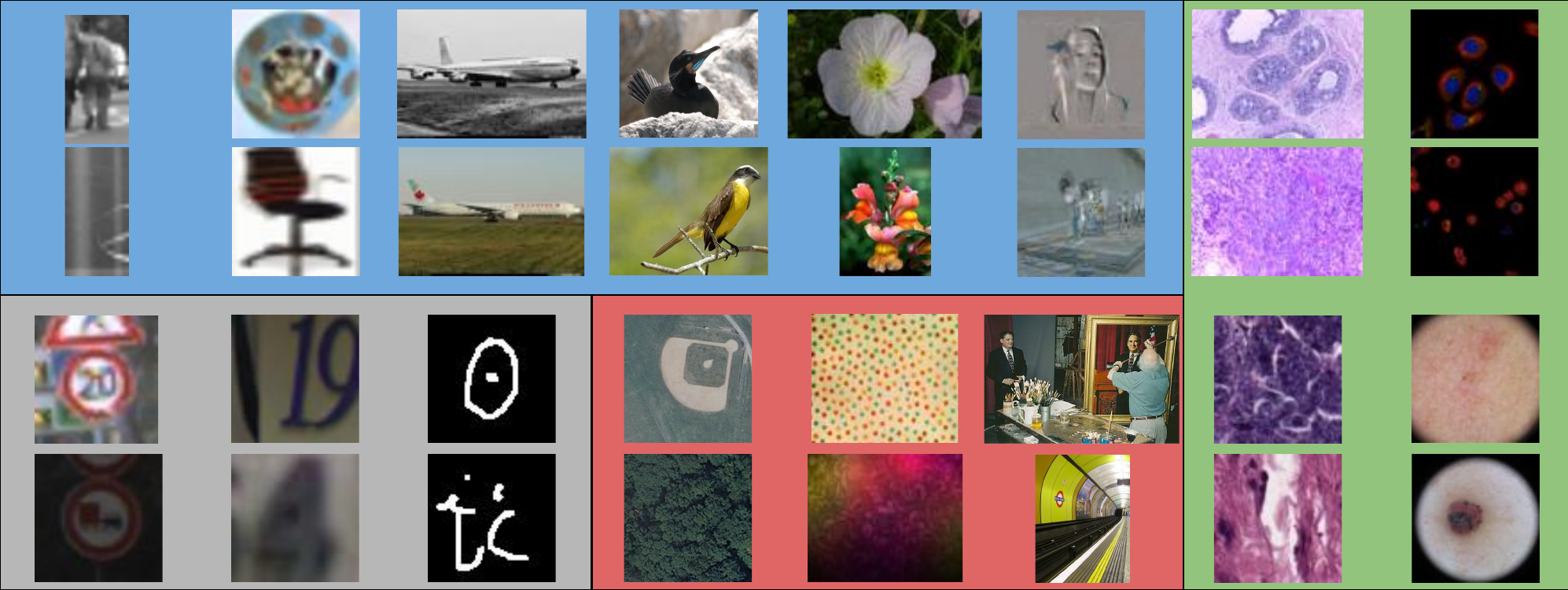}
   \caption{
Samples from all datasets. 
Top rows: Daimler Pedestrians, CIFAR100, FGVC-Aircraft, CU Birds, VGG-Flowers, UCF101, BACH, Protein Atlas.
Bottom rows: GTSRB, SVHN, Omniglot, UC Merced Land Use, Describable Textures, Indoor Scenes, Kather, ISIC. 
Color coding is by group: \textcolor{blue}{\Natural} \textcolor{mygray}{\Symbolic} \textcolor{red}{\Scenes} \textcolor{green}{\Biological}
}
\label{fig:datasets}
\end{figure}

We include 16 datasets in our experiments, significantly more than all prior work.
Dataset samples are shown in Figure~\ref{fig:datasets}.
We group these datasets into 4 categories: \textbf{\Natural}, \textbf{\Symbolic}, \textbf{\Scenes}, and \textbf{\Biological}.
A summary is shown in Table~\ref{table:datasets}.
Some of the datasets in the first three groups are also in the Visual Domain Decathlon (VDD)\cite{vdc1}, a multi-task learning benchmark.

\begin{table}[!tbh]
\centering
\caption{
Summary of the 16 datasets included in our experiments: encompassing fine-grain, symbolic, scene, textural, and biological domains.
This is the first work exploring self-supervised representation learning on almost all of these tasks
}
\resizebox{0.99\linewidth}{!}{
\begin{tabular}{ c l  c c c c}
\toprule
 \textbf{Name} & Type & Size (Train) & Coarse/Fine & Abbreviation  \\
\midrule
Daimler Pedestrians\cite{pedestrians} & Road Object & 20k & Coarse & PED \\
CIFAR100\cite{cifar} & Internet Object & 40k & Coarse & C100 \\
FGVC-Aircraft\cite{aircraft} & Internet Object & 3.3k & Fine & AIR\\
Caltech-UCSD Birds\cite{cub} & Internet Object & 8.3k & Fine& CUB \\ 
VGG-Flowers\cite{flowers} & Internet Object & 1k & Fine & FLO \\ 
UCF101\cite{ucf1,ucf2} & Pseudo-Internet Action & 9.3k & Coarse & UCF \\
\midrule
German Traffic Signs \cite{gtsrb} & Symbolic & 21k & Coarse & GTS \\
Street View House Numbers\cite{svhn} & Symbolic & 59k & Coarse  & SVHN\\
Omniglot\cite{omniglot} & Symbolic & 19k & Fine & OMN \\
\midrule
UC Merced Land Use\cite{merced}& Aerial Scene & 1.5k & Coarse & MER \\
Describable Textures\cite{textures} & Texture & 1.9k & Fine & DTD \\
Indoor Scene Recognition \cite{scenes}& Natural Scene & 11k & Coarse & SCE \\
\midrule
ICIAR BACH\cite{bach} & Biological & 240 & Coarse  & BACH \\
Kather\cite{kather} & Biological & 3k & Coarse & KATH\\
Protein Atlas\cite{protein_atlas} & Biological & 9k & Fine & PA \\
ISIC\cite{isic1,isic2} & Biological & 17k & Coarse & ISIC \\
\end{tabular}
}
\label{table:datasets}
\end{table}

\textbf{\Natural\ Object Recognition:}
This group consists of object recognition problems on internet imagery.
We include both coarse-grained (CIFAR100, Daimler Pedestrians) and fine-grained (FGVC-Aircraft, CUB, VGG Flowers) object classification tasks.
Finally, we  include the ``dynamic images'' of UCF101, a dataset that possesses many of the same qualitative attributes of the group.

\textbf{\Symbolic :}
We include three well-known symbolic tasks: Omniglot, German Traffic Signs (GTSRB), and Street View House Numbers (SVHN). 
Though the classification problems might be deemed simple, these offer domains where classification is very different from natural internet imagery: texture is not generally a useful cue and classes follow strict explainable rules.

\textbf{\Scenes :}
These domains, UC Merced Land Use (satellite imagery), Describable Textures, and Indoor Scenes, all require holistic understandings, none having an overarching definition of object/symbol. 
Indoor Scenes does contain internet imagery as in our first group, but is not object-focused. 

\textbf{\Biological :}
BACH and Kather consist of histological (microscopic tissue) images of breast and colon cancer respectively, with the classes being the condition/type of cancer.
Protein Atlas is microscopy images of human cells, with the goal being classification of the cell part/structure shown.
Finally, ISIC is a dermatology dataset consisting of photographs of different types of skin lesions.

\begin{figure}[!t]
\centering
\includegraphics[width=\linewidth]{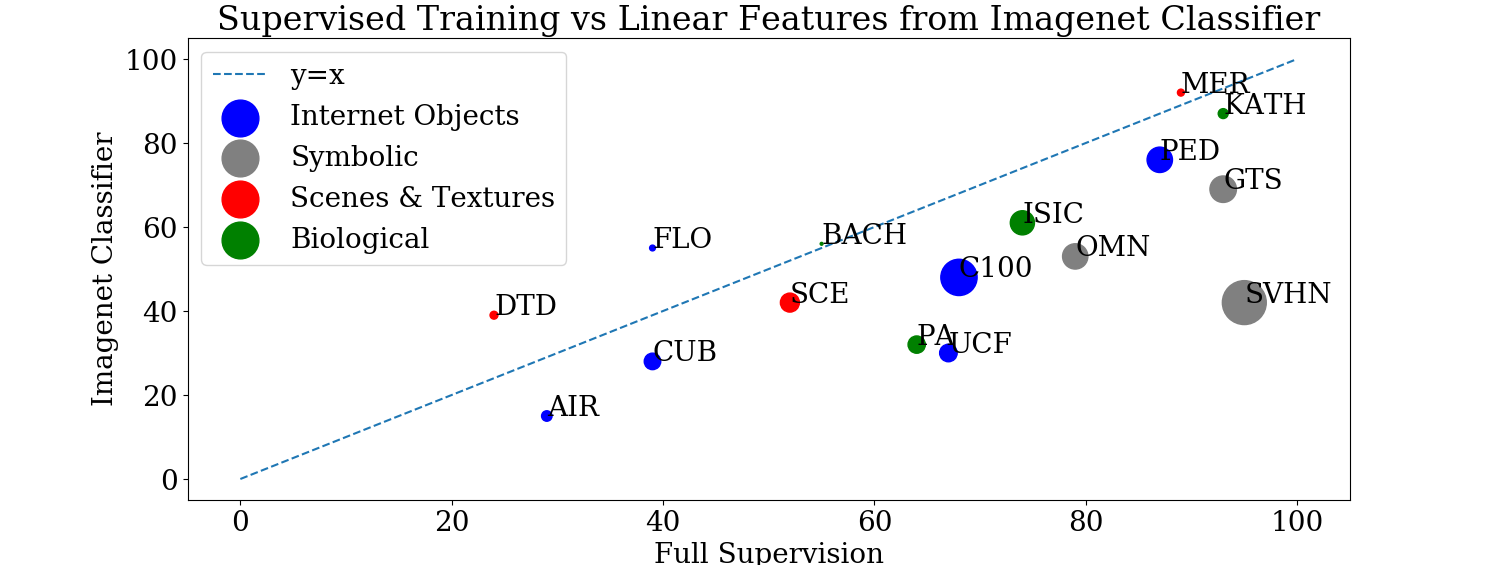}
   \caption{
Test accuracy of a fully supervised network vs. a linear classifier on top of an ImageNet-classification frozen feature extractor.
Marker area indicates dataset size
}
\label{fig:sup_vs_imagenet}
\end{figure}

Before evaluating self-supervision, we study the datasets themselves in terms of difficulty and similarity to ImageNet.
To do so, we compare the accuracy of a network trained from scratch with that of a linear classifier operating on an ImageNet-pretrained network (Figure~\ref{fig:sup_vs_imagenet}).
The higher of the two numbers measures the difficulty, while their relationship quantifies the similarity to ImageNet.

We find that small datasets in the \Natural\ domain tend to be the hardest, while large \Symbolic\ datasets are the simplest.
The symbolic tasks also have the largest gap between supervision and feature extraction, suggesting that these are the farthest from ImageNet.
Overall, the ImageNet feature extractor performance is strongly linearly correlated to that of the fully supervised model ($p=0.004$).
This is expected for the \Natural\ domain, but the similar utility of the feature extractor for the \Biological\ domains is surprising.
Dataset size also plays a role, with the pretrained feature extractor working well for smaller datasets.

For fine-grained classification (AIR, CUB, FLO), the supervised models perform comparably, while the ImageNet classifier's performance varies widely.
In addition to the small size of VGG-Flowers, this case is also partly explainable by how these datasets overlap with ImageNet.
Almost half of the classes in ImageNet are animals or plants, including flowers, making pretraining especially useful.

\section{Methods}\label{sec:methods}

\subsection{Self-Supervised Learning Techniques}

In this paper we look at three popular methods,  Rotation, Jigsaw, and Instance Discrimination.
We also look at the classical technique of Autoencoders as a baseline for the large variety of autoencoder-based pretexts\cite{split_ae,colorization,context_ae}.
We briefly describe each method below, please view the cited works for detailed information.

\textbf{Learning by Rotation:}
A network is trained to classify the angle of a rotated image among the four choices of 0, 90, 180, or 270 degrees\cite{rotation}.

\textbf{Learning by Solving Jigsaw Puzzles:}
The image is separated into a 3x3 grid of patches, which are then permuted and fed through a siamese network which must identify the original layout\cite{jigsaw,siamese}.
We use 2000 training permutations in our experiments, finding this offered superior performance to 100 or 10,000 under our hyperparameters.

\textbf{Instance Discrimination:}
Instance Discrimination (ID) maps images to features on the unit sphere with each image being considered as a separate class under a non-parametric softmax classifier\cite{inst_disc}.

\textbf{Autoencoders:}
Autoencoders were one of the earliest methods of self-supervised learning\cite{split_ae,denoising_ae,context_ae,bengio_ae,hinton_ae}.
An autoencoder learns an encoder-decoder pair of networks that reconstruct the input image.

\subsection{Architecture \& Evaluation}
We resize inputs to $64 \times 64$ and use a ResNet26, as in Rebuffi et al. \cite{vdc1,vdc2}. 
The lower resolution eases computational burden as well as comparison and future adaptation to the VDD.
For Autoencoding, a simple convolutional decoder is used.
Features maps of size $256\times 8 \times 8$ are extracted before the final pooling layer and average pooled to 256, 4096 ($256 \times 4 \times 4$), or 9216 ($256 \times 6 \times 6$), with 256 being the default.
A linear classifier is trained on these features.
Training/architecture details are in the Supplementary.


\section{Related Work} 

There are three pertinent recent surveys of self-supervision. 
The first is by Kolesnikov et al. who evaluate several methods on ImageNet and Places205 classification across network architectures\cite{google_survey}.
We focus on many domains, an order of magnitude more than their work.
The second relevant survey is by Goyal et al., who introduce a benchmark suite of tasks on which to test models as feature extractors.
While they scale on a variety of downstream tasks, the pretraining datasets are all internet imagery, either ImageNet or YFCC variants\cite{fair_benchmark,YFCC}.
Our work includes a much wider variety of both pretraining and downstream datasets.
VTAB tests \textit{pretrained feature extractors} on a variety of datasets, performing self-supervised learning only on ImageNet\cite{vtab}.
Finally, a concurrent paper evaluates these self-supervised techniques as an auxilliary loss for few-shot learning\cite{selfsupfewshot}.

One trend of inquiry concerns classifiers that perform well on multiple domains while sharing most of the parameters across datasets\cite{vdc1,vdc2,uodb}.
The pre-eminent examples of this are by Rebuffi et al. \cite{vdc1,vdc2} who present approaches on the VDD across 10 different domains.
We use these datasets (and more) in our training, but evaluate \textit{self-supervised approaches} in \textit{single-domain} settings.

There has also been prior work using problem/domain-specific SSRL methods, such as \cite{biomed1,biomed2,biomed3,biomed4} in the biological and medical fields or \cite{satellite} for aerial imagery.
Many of these approaches use variations of autoencoding as the pretext task; we include autoencoding in our evaluation.
In contrast to these, our focus is on the cross-dataset applicability of these pretexts.



Other SSRL methods include generative models\cite{donahue_generative,dumoulin_generative,goodfellow_generative}, colorization \cite{larsson_colorization,colorization}, video-based techniques\cite{owens_video,desa_multimodal,pathak_video,misra_video,goroshin_video,walker_video}, or generic techiques\cite{deep_cluster,noise_as_targets,moco,pirl,simclr}.
It is very possible that a subset of these methods could offer improved performance on some of the domains we work with, however in this work we focus on the popular fundamental methods of Rotation, Jigsaw, and Instance Discrimination as well as Autoencoding as a representative set.
Doersch and Zisserman use multiple pretexts simultaneously to improve representation quality \cite{doersch_multitask}.
This approach could be complementary to the work featured here.
Semi-supervised approaches such as S4L\cite{s4l} are relevant to Section~\ref{sec:semi}.

\section{Downstream task Performance Analysis}

\subsection{Downstream task accuracy}\label{sec:rankings}

\begin{figure}[!tbh]
\centering
\includegraphics[width=\linewidth]{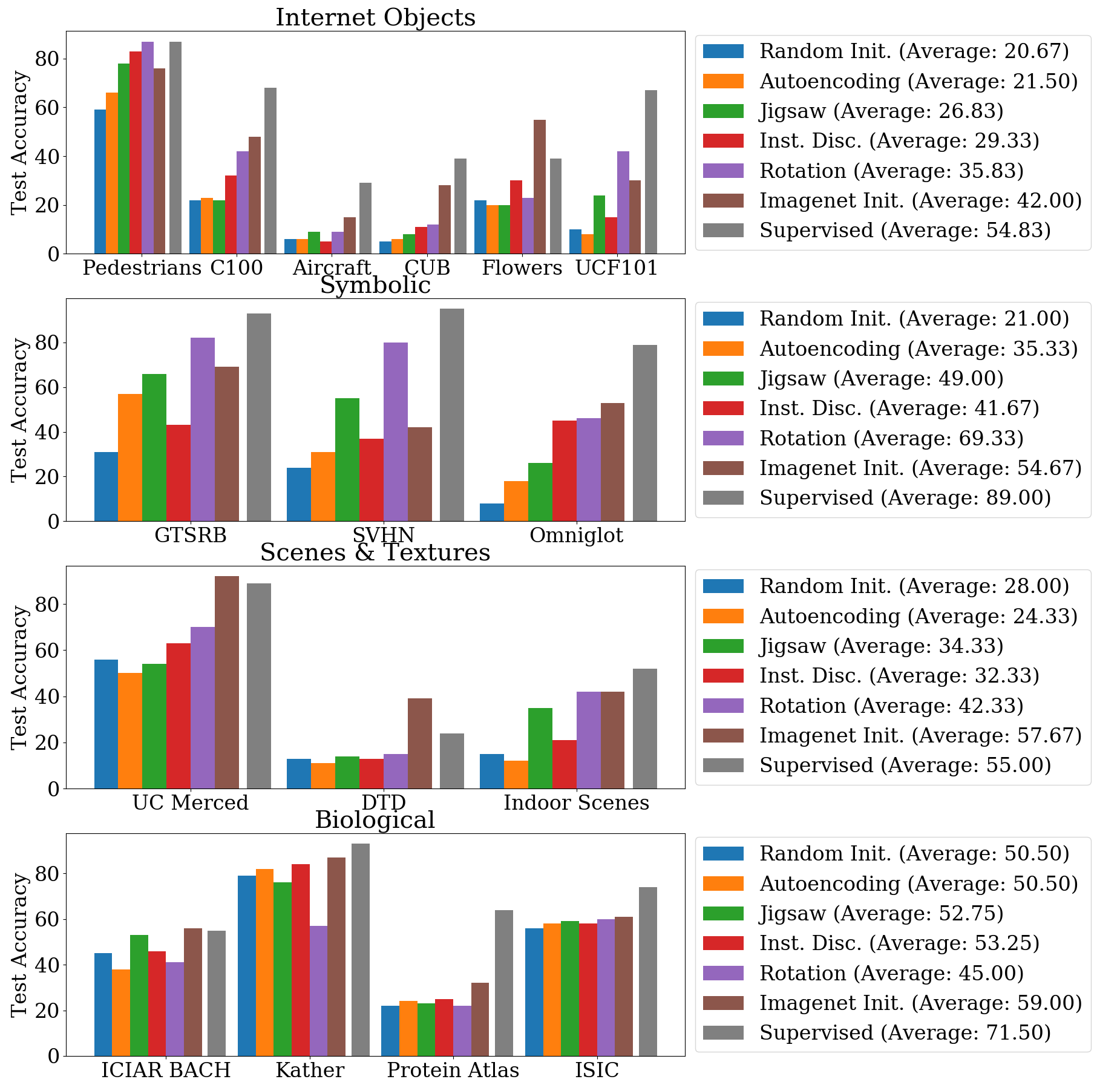}
   \caption{
Downstream classification accuracies for each pretext, as well as a randomly initialized feature extractor, a frozen ImageNet classifier, and fully supervised training.
Rotation achieves the highest accuracies on the first three groups, but fails on the Biological domain, where Instance Discrimination performs best.
The relative rankings of Jigsaw and ID vary between groups and are practically equal in overall average
}
\label{fig:result_summary}
\end{figure}

A summary of downstream testing accuracies is shown in Figure~\ref{fig:result_summary}. 
The obvious question to ask in this investigation is ``which pretext is best"?
On ImageNet, per the respective original works, the ordering (best to worst) was Rotation, Instance Discrimination, Jigsaw, Autoencoding.
We see that this ranking is not universal: while Rotation is the best pretext on the first three groups, it lags behind even random initialization on the \Biological\ domains.
Furthermore, the relative rankings of ID and Jigsaw vary between groups.
We investigate what powers the performance of each of these methods in Section~\ref{sec:success_method}.

\subsubsection{Limited Label Training}\label{sec:semi}

\begin{figure}[!tb]
\centering
\includegraphics[width=\linewidth]{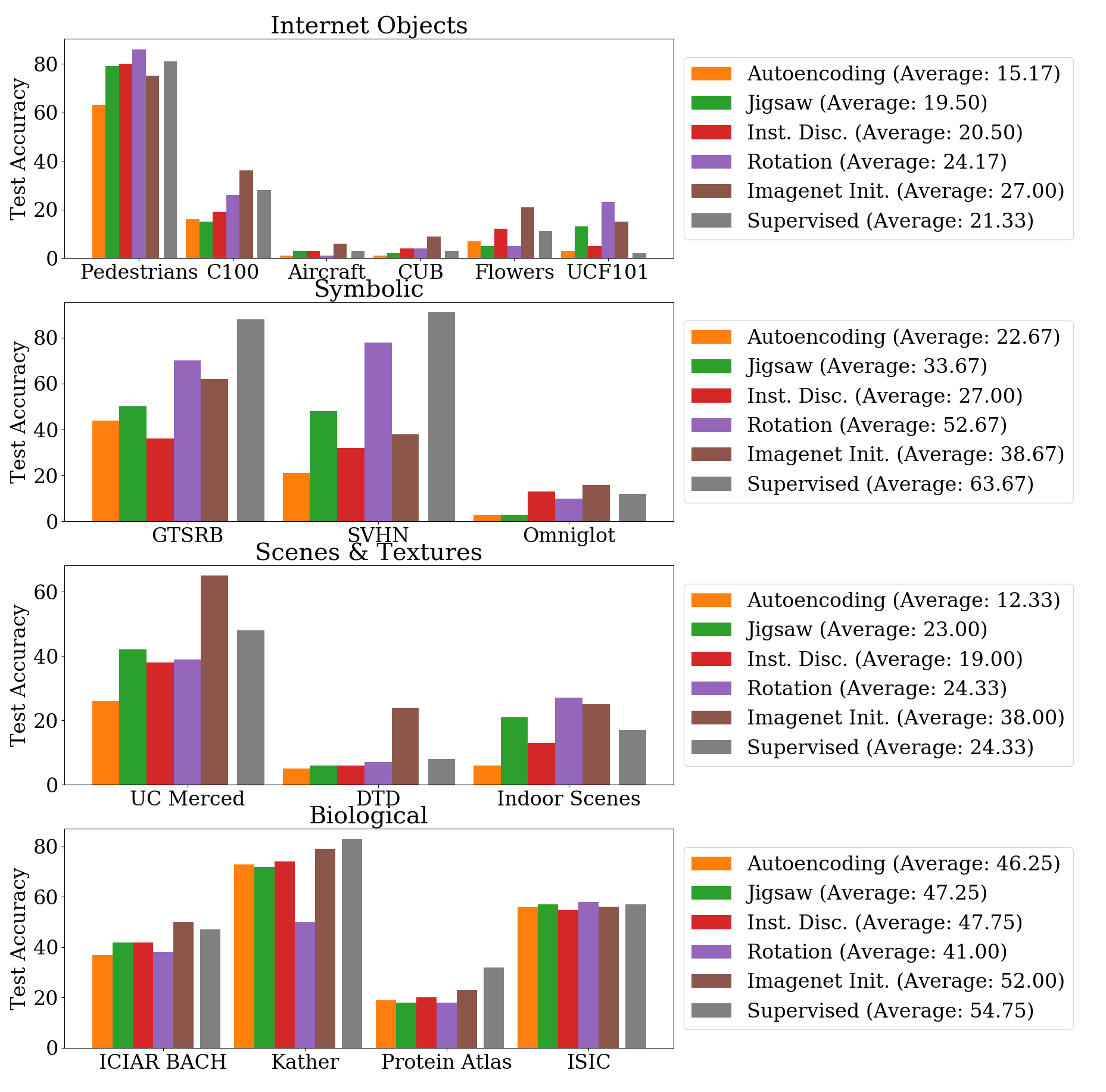}
   \caption{
Pretext feature extractors trained on the entire dataset, then a linear head trained on 10\% of the labels.
The fully supervised method simply trains on 10\% of the dataset.
We observe Rotation matching/outperforming Supervision for \Natural\ and \Scenes\ imagery.
Autoencoding performs very well on the Biological domains in the semi-supervised setting, a novel result that indicates the potential for future development (given that this Autoencoder implementation is the simplest possible)
}
\label{fig:semi_sup_grouped_bar}
\end{figure}

Self-supervised learning has made an impact in the field of \textit{semi}-supervised learning, where only a subset of the dataset given is labeled, as in the work of Zhai et al.\cite{s4l}.
In that work on ImageNet, a self-supervised feature extractor followed by a supervised linear head falls well short of the purely supervised baseline (40\% accuracy compared to 80\% when 10\% of labels are available).
We find that this conclusion does not hold on our domains when 10\% of labels are used, with a pretext + linear setup outperforming the fully supervised models on some datasets/groups (Figure~\ref{fig:semi_sup_grouped_bar}).\footnote{ Note that \cite{s4l} performs extensive hyperparameter tuning for the supervised baseline.}

On the \Natural\ and \Scenes\ groups, Rotation matches/outperforms full supervision.
On ISIC, both Instance Discrimination and Jigsaw match Supervision.
Interestingly, Autoencoding performs well in the \Biological\ domains.
Given the difficulty of expert annotation in the medical field, this is a valuable finding to encourage Autoencoder-based methods in these label-scarce domains, vindicating choices in past work\cite{biomed1,biomed2,biomed3,biomed4}.
The unlabeled 90\% of the data being available for SSRL methods is critical: when only the labeled subset is used for pretraining the performance drops an average of approximately 10\%.

\subsection{Inspecting Failure Modes}\label{sec:failure}
We call specific attention to the problems where self-supervised techniques do not achieve even \emph{half} of the supervised accuracy (Aircraft, CUB,  Textures and Protein Atlas).
These seem to involve two kinds of problems.


\subsubsection{Textures and Protein Atlas:}\label{sec:failure_scenes_bio}
In both these datasets, the entities being classified are not objects of recognizable shape, which is true of most \emph{object recognition} datasets where the self-supervised techniques were developed.
The images also do not have a canonical orientation, unlike in internet imagery where gravity and photographers' biases provide such an orientation.
We hypothesize that the lack of orientability hobbles Rotation, and the textural nature of both problems results in there being little to distinguish a patch in the Jigsaw pretext from a complete image, meaning Jigsaw can do little besides cue off of low-level properties (such as chromatic aberrations~\cite{jigsaw})
As such, modulo low-level dataset biases, the Rotation and Jigsaw pretext tasks \emph{cannot even be solved in these domains}.

\subsubsection{Fine-Grained Classification on Aircraft and CUB:}\label{sec:fine_fail}
\begin{figure}[!tb]
\centering
\includegraphics[width=\linewidth]{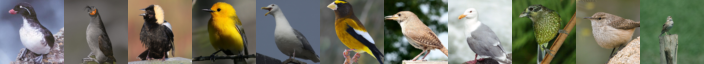}
   \caption{
The 10 nearest neighbors (in order) of the far-left image in the CUB validation set measure by the unpooled feature space of the Rotation pretext model.
We see that coloring plays almost no role in determining neighbors, but pose is a very strong signal
}
\label{fig:CUB_knn}
\end{figure}

In these datasets, the Rotation and Jigsaw pretexts are solvable, involving objects captured in a canonical orientation.
Even so, neither pretext learns a good representation.
While the \emph{domains} are favorable to the pretexts, the \emph{tasks} are not: both tasks involve fine-grained distinctions in color or texture, and subtle differences in shape. 
One hypothesis is that modeling these subtle differences is not necessary for solving the pretext task, causing the learnt representation to be \emph{invariant} to these vital distinctions.
Indeed, when we look at nearest neighbors in CUB for our Rotation model, we see birds of completely different colors but in the same pose, suggesting that the representation has captured pose, but ignored color (Figure~\ref{fig:CUB_knn}).
We quantitatively evaluate this hypothesis further in Section~\ref{sec:success_method}.

Note that we have not discussed the failures of Instance Discrimination in both domains.
Instance Discrimination must learn to distinguish between individual images, which should be solvable in any domain as long as the images are distinct.
As such, its failure and general performance is much more unpredictable and less interpretable, as we elaborate on in the next section.

\subsection{Reasons for Success}\label{sec:success_method}

The flip side of why they fail is why they succeed.
In prior work on the domain of internet imagery, this has mostly been answered by intuition.
Rotation and Jigsaw were engineered as tasks that ostensibly require semantic understanding to solve.
Instance Discrimination's success is similarly intuitive: spreading points in a constrained space will naturally cluster similar images.
Given the failures above, neither of these intuitions endure without modification in the varied domains of our work, and a more nuanced reasoning is necessary.

\begin{figure}
\centering
\includegraphics[width=0.65\linewidth]{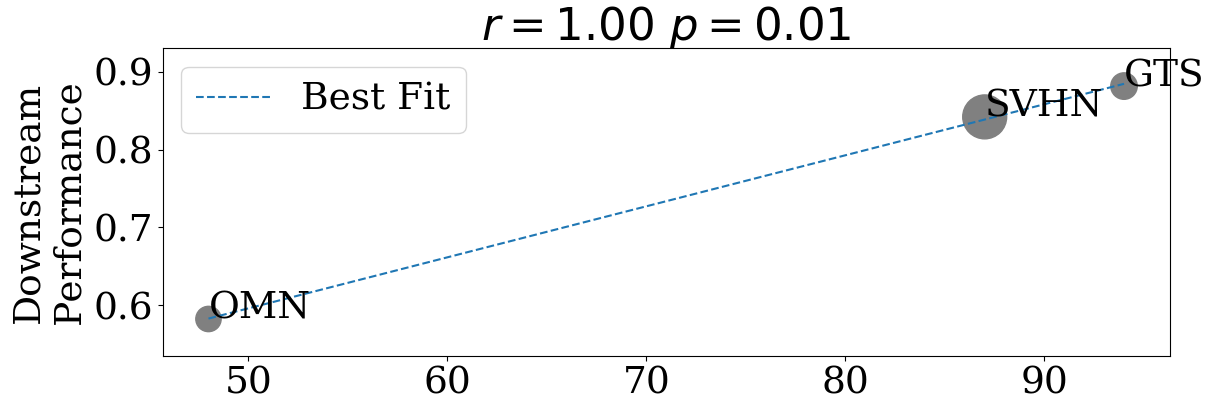}
   \caption{
Rotation pretext testing accuracy vs. unnormalized downstream performance.
Almost perfect correlation is evident
}
\label{fig:rotation_symbolic}
\end{figure}

\paragraph{Semantic understanding:}
As discussed above, for some domains such as Aircraft or CUB, the pretexts do not produce a good semantic representation perhaps because they do not require a semantic understanding to solve.
As an additional example, Jigsaw classifies permutations on Omniglot with 77\% accuracy, but performs poorly on the classification task: line-matching is not understanding.
Another example is Kather, where Rotation picks up on some hidden cue to fully solve the pretext problem without attaining any semantic knowledge.
On the other side of the spectrum, we observe on the \Symbolic\ domains in Figure~\ref{fig:rotation_symbolic} that Rotation \textit{is} implicitly performing semantic classification, making it a near-optimal pretext task.
\emph{When} is semantic understanding a prerequisite for a pretext task, and how can we test this?


\begin{figure}[t]
\centering
\includegraphics[width=\linewidth]{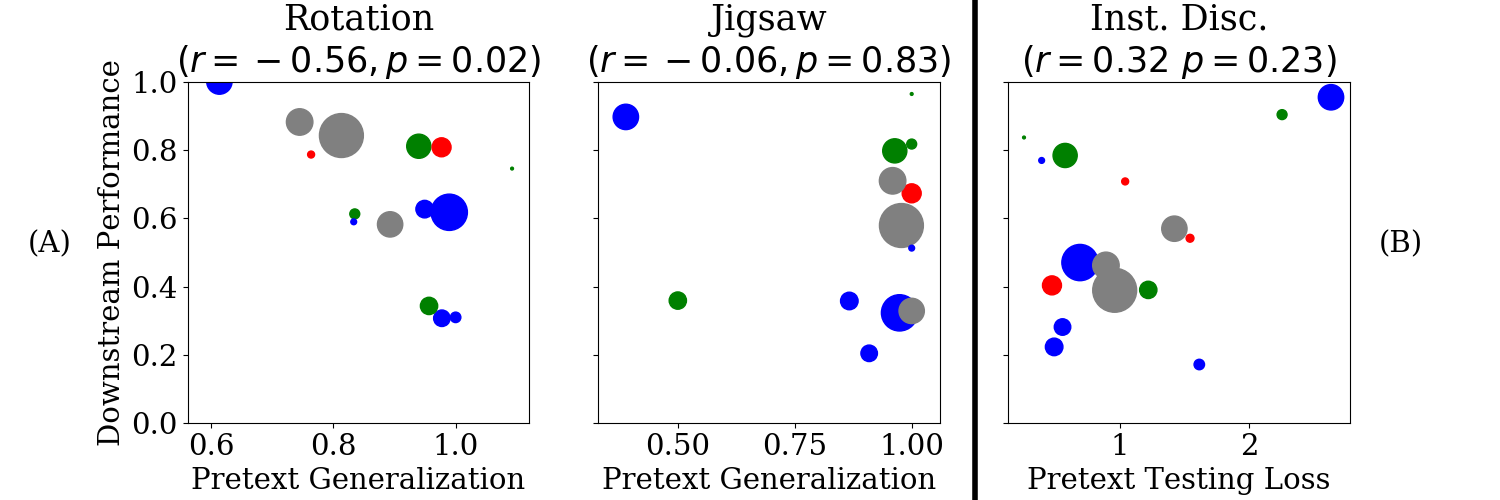}
   \caption{
\textbf{(A)}
Generalization of the pretexts to novel classes is plotted on the x-axis, calculated as $\frac{PretextAcc_{Test}}{PretextAcc_{Val0.5}}$, where Train/Val0.5 contain only half the classes of their usual counterparts.
On the y-axis is downstream accuracy normalized by Supervision accuracy \textit{for the regularly trained model}, e.g. $\frac{ClassAcc_{Rot}}{ClassAcc_{Sup}}$.
We see a strong correlation for Rotation, but none for Jigsaw.
\textbf{(B)}
Instance Discrimination pretext loss vs. downstream classification accuracy, no significant correlations.
Marker area indicates dataset size and color the domain grouping.
}
\label{fig:half_classes_and_id_downstream}
\end{figure}

We propose that semantic understanding is a prerequisite for a pretext \textit{when the solution method is class-dependent}: in this case, the network must necessarily implicitly classify the image to perform the pretext well.
For example, in the fine-grained classification of airplanes and birds, determining the orientation of the object is class-\emph{in}dependent, so a network trained on the Rotation pretext does not need to learn features indicative of class.
One way of testing if the pretext solution is class independent is to see if a network trained to solve the pretext on one set of classes solves the pretext on \emph{unseen} classes.
We call this test \emph{pretext generalization} (Figure~\ref{fig:half_classes_and_id_downstream}): we train and validate pretexts on only half of the available classes, and then evaluate the pretext on the entire test set.
We predict that worse pretext generalization should imply better downstream performance.


Our prediction is correct for Rotation, affirming our hypothesis, but Jigsaw models seem to show high pretext generalization in almost all cases.
This lack of correlation might be because Jigsaw is relying on low-level cues that do generalize \emph{in addition to} the semantic information.
We contrast these findings with a concurrent paper\cite{selfsupfewshot} which speculates that Rotation is not as useful for few-shot learning on FGVC-Aircraft or VGG-Flowers due to the relative difficulty of the pretext task.

For Instance Discrimination, counterintuitively, downstream accuracy is \emph{not} correlated with pretext loss (Figure~\ref{fig:half_classes_and_id_downstream}B).
Our proposal is thus not the complete picture.

\begin{figure}[!t]
\centering
\includegraphics[width=\linewidth]{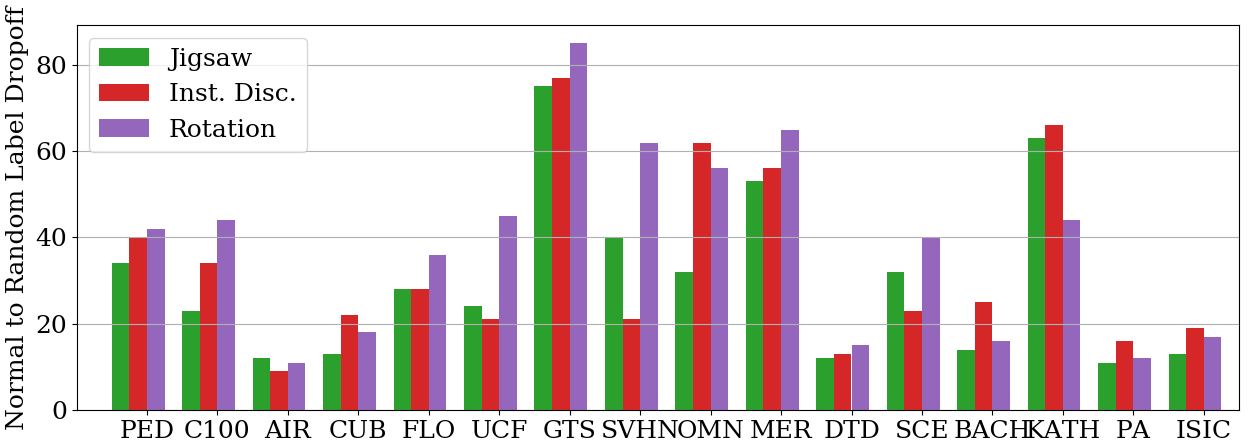}
   \caption{
Difference between normal and random label training accuracy.
Note that even though Instance Discrimination had lower average \textit{validation} accuracy than Rotation across all datasets, they actually had similar (normal label) average \textit{training} accuracy at 56\% (Jigsaw had 47\%).
}
\label{fig:bar_random_diffs}
\end{figure}

\paragraph{Linear separability:}
The above discussion assumes that the relationship between the pretext task and the downstream task is the key.
But what if the downstream task is immaterial? 
Pretext tasks might be succeeding by simply enabling \emph{all} tasks, by creating a feature space where \emph{many} labelings of the data points are expressible as linear classifiers.
To test this hypothesis, we retrain linear classifiers with randomly shuffled training labels and compare this (training) accuracy to the training accuracy with correct labels. 
We find that these quantities are more correlated for Jigsaw than Instance Discrimination or Rotation (Table~\ref{table:random_v_normal}), per task decreased are shown in Figure~\ref{fig:bar_random_diffs}.
This means that Jigsaw succeeds more by learning generic feature descriptors of images rather than by capturing semantics specific to the downstream task, while this is less true for Rotation or Instance Discrimination.
Note that here we are talking of the training accuracy, or the empirical risk; this experiment does not reveal how or why pretext methods generalize to data not in the training set. 

\begin{table}[!t]
\centering
\caption{
Pearson correlations and p-values for training accuracy with random labels vs. normal.
Jigsaw and ID have significant correlations, while Rotation trails substantially
}
\ra{1.3}
\begin{tabular}{@{}lccccccccc@{}}\toprule
&& \multicolumn{2}{c}{Jigsaw} & \phantom{abc}& \multicolumn{2}{c}{Inst. Disc.} &
\phantom{abc} & \multicolumn{2}{c}{Rotation.}\\
\cmidrule{3-4} \cmidrule{6-7} \cmidrule{9-10}
$(r,p)$  && 0.59 & 0.02 && 0.46 & 0.07  && 0.45 & 0.08\\ 
\bottomrule
\end{tabular}
\label{table:random_v_normal}
\end{table}

\section{Feature Space Exploration}
The discussion above suggests the virtue of analyzing the learnt representations independent of the downstream task.
Below, we look at the \emph{intrinsic dimensionality} of the learned representations, and the resulting notions of similarity.

\subsection{Implicit Dimensionality of the Representations}\label{sec:pca}
\begin{figure}[!t]
\centering
\includegraphics[width=\linewidth]{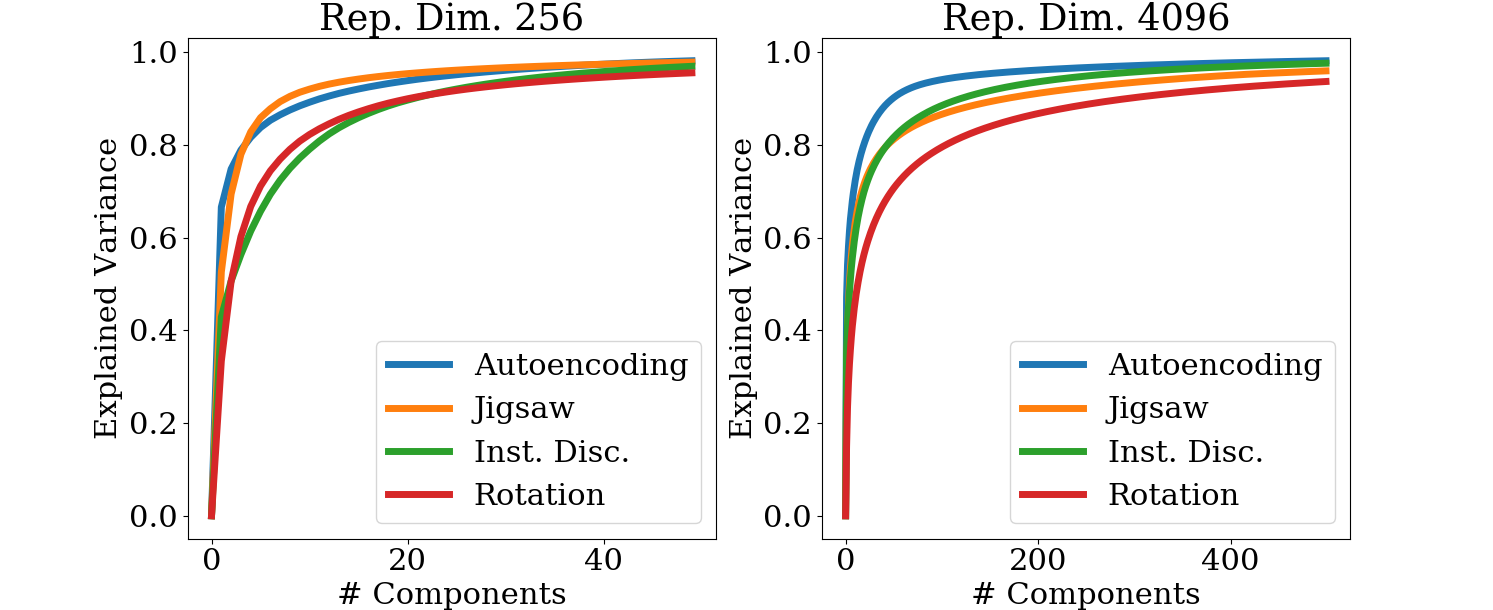}
   \caption{
Plot of the average fraction of explained variance vs. number of principal components (validation sets); a higher number means greater explanation, and thus relatively \textit{lower} implicit dimension.
Bach is omitted due to its extremely small size
}
\label{fig:summed_pca}
\end{figure}

\begin{figure}[!t]
\centering
\includegraphics[width=\linewidth]{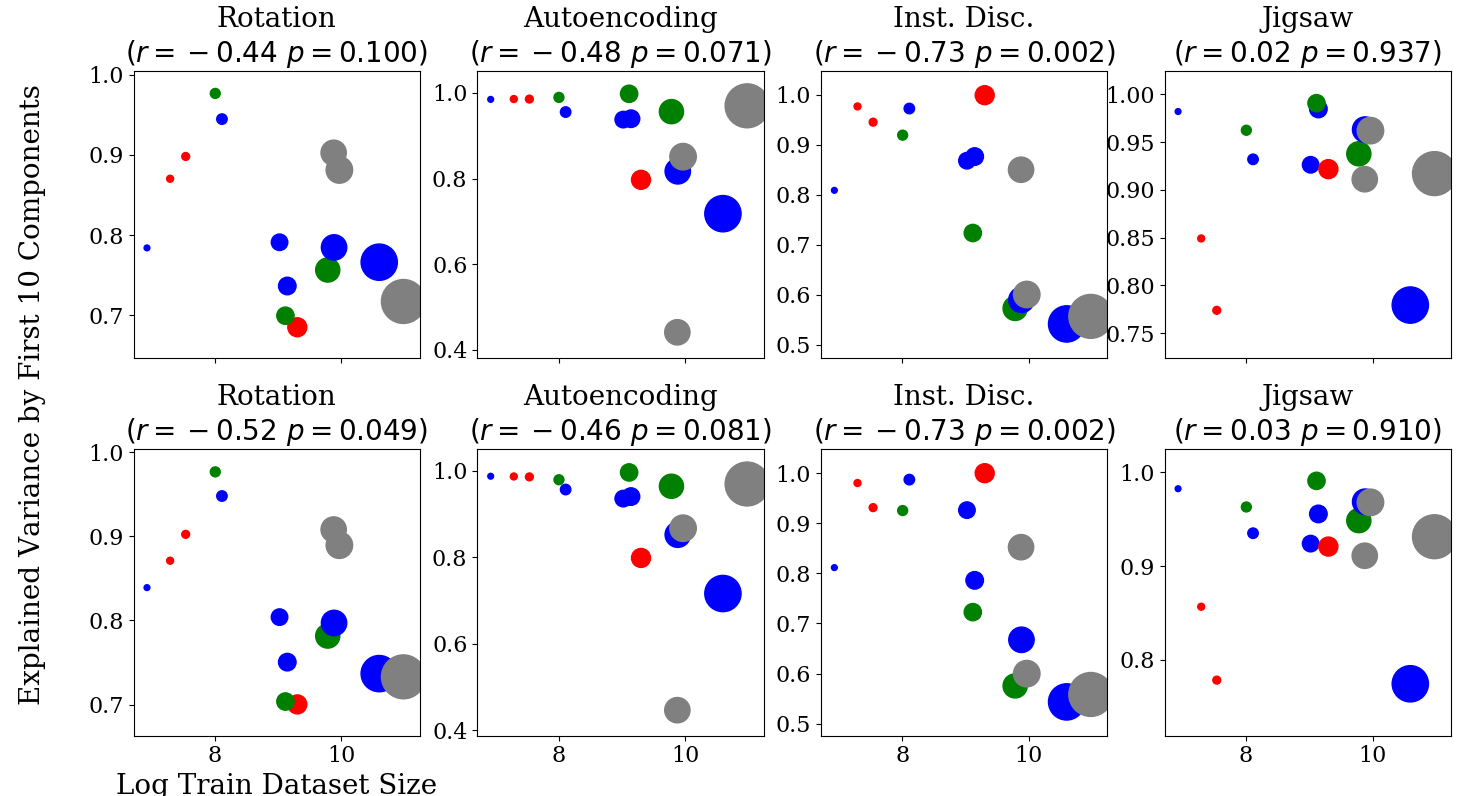}
   \caption{
The fraction of variance explained by the first 10 components vs. the log of the training set size.
Top row is training PCA, bottom row is validation PCA.
Moderate to strong correlations exists for all pretext tasks besides Jigsaw.
Marker area indicates dataset size and the colors are domain groupings.
}
\label{fig:pca_dim_vs_log_size}
\end{figure}

The dimensionality of the representation is largely regarded as a hyperparameter to be tuned.
Kolesnikov et al. show that a larger representation is always more useful when operating on large datasets, but this comes with a tradeoff of memory/storage\cite{google_survey}.
What has not been studied is the \textit{implicit} dimensionality of the representations, such as via Principal Component Analyis (PCA).
Intuitively, one would expect the 4-way Rotation task to produce compact representations, with Instance Discrimination using all available dimensions to spread out the points as the loss demands.
We find that this is \textit{not} the case, via performing PCA on the representations and summing the explained variance of the first $n$ values averaged across all datasets (Figure~\ref{fig:summed_pca}).
We also find that the implicit dimensionality induced by each pretext varies considerably across domains (Figure~\ref{fig:pca_dim_vs_log_size}).

Surprisingly, Instance Discrimination and Rotation have extremely similar implicit dimensionality in the 256-dimensional case. 
Also note that Instance Discrimination clearly is not fully utilizing the available latent space, as the first 40 components explain over 95\% of the variance.
In the higher-dimensional example, Rotation has by far the largest implicit dimensionality. 

We next investigate the effect of dataset size on implicit dimensionality.
Intuition says that a larger dataset would demand a more expressive representation on every task, and we see in Figure~\ref{fig:pca_dim_vs_log_size} that this indeed holds true for \textit{all} tasks except for Jigsaw.
We hypothesize the lack of dimensionality increase for Jigsaw is because it exploits relatively low-level attributes instead of semantic knowledge. 

\subsection{Nearest Neighbors}\label{sec:neighbors}

As seen in Figure~\ref{fig:CUB_knn}, the nearest neighbors in feature space can yield great insight into the inner workings of our models.
We repeat this experiment for all of our main methods on three datasets (CUB, Omniglot, Kather) in Figure~\ref{fig:nearest_neighbors} (additional examples in the supplementary).

\begin{figure}[!tbh]
\centering
\includegraphics[width=\textwidth]{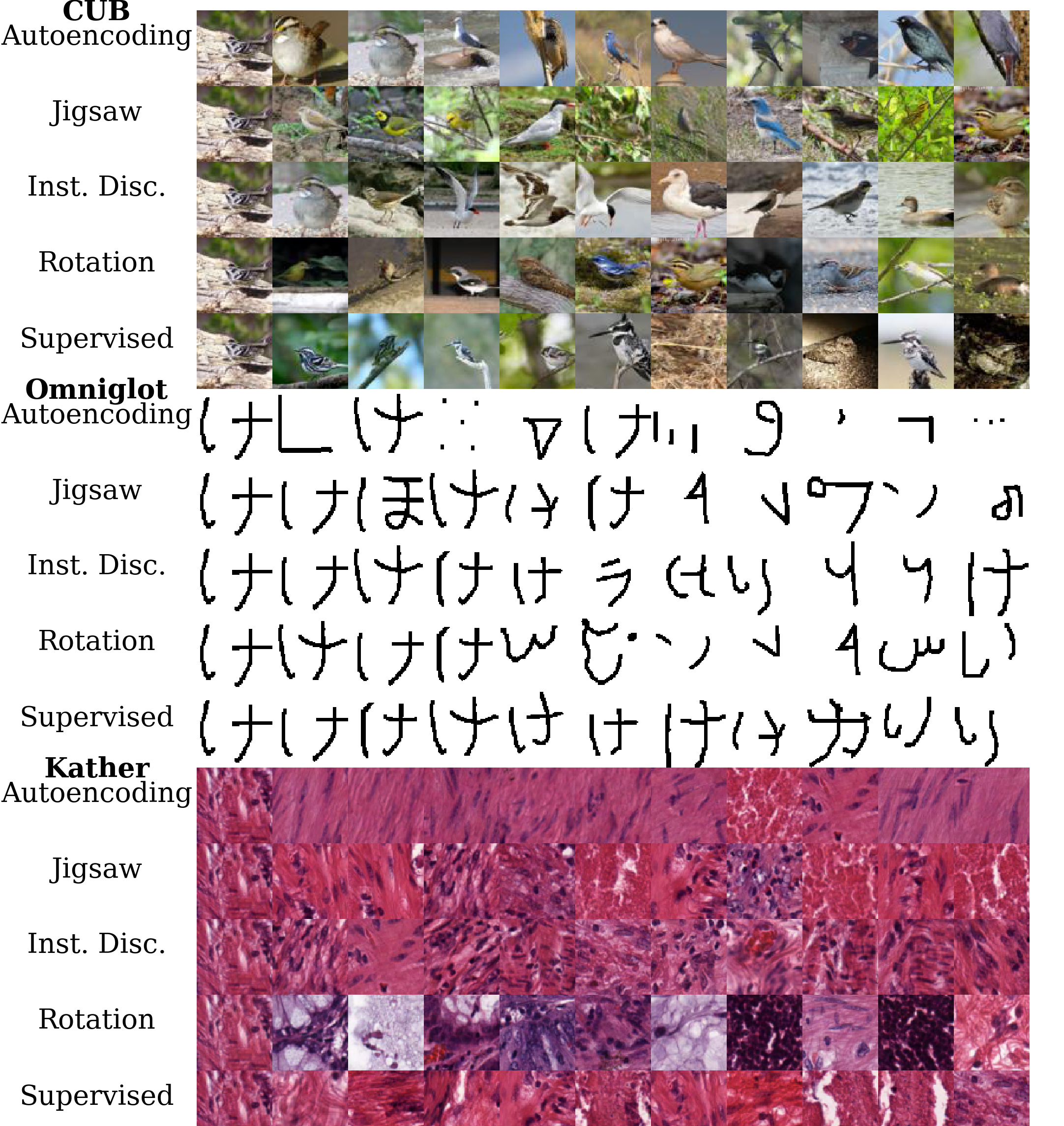}
   \caption{
Each block of images comes from a single dataset, with each row corresponding to the pretext method.
The farthest left image is the base image from which the distance in feature space is calculated to all the others in the validation set.
Left-to-right is increasing distance in feature space among the 10 nearest neighbors
}
\label{fig:nearest_neighbors}
\end{figure}


\textbf{CUB:}
The strong pose-capturing of Rotation is again observed. 
Jigsaw exhibits similar phenomenon slightly more weakly, and also favors similarly cluttered edge-heavy backgrounds as in the base image.
Instance Discrimination appears to have reasonably matched the species in the first and tenth neighbor, but in between has many unrelated birds.


\textbf{Omniglot:}
The poor performance of Autoencoding is reflected in its matching ability.
Rotation exhibits strong retrieval, with Jigsaw offering similar but weaker results.
Instance Discrimination is a similar case to CUB, where a discernable trend is hard to spot, but several correct characters are matched.

\textbf{Kather:}
No method improved much on random initialization (79\% accuracy).
Reflectively, Autoencoding, Jigsaw, and ID all simply match reasonably visually similar results.
Rotation, however, failed catastrophically (\textless60\% accuracy), despite high pretext accuracy (95\% testing).
We hypothesized that the pretext was exploiting a spurious cue, which is affirmed by radically varying neighbors.

\section{Conclusion}
In this work, we have for the first time explored the notion of self-supervised learning in small-scale images in novel domains, identifying three domains where all current self-supervised methods have need of development: fine-grained, textural, and biological domains.
In addition, we have revealed intriguing properties of the pretexts and the corresponding learnt representations, whose impact deserves further study.
We hope that the release of our codes, models, and formatted dataset splits will help aid progress on all of these fronts.


\textbf{Acknowledgements:} This work was funded by a DARPA LwLL grant.

{\small
\bibliographystyle{splncs04}
\bibliography{egbib}
}

\end{document}


\pagestyle{headings}
\mainmatter
\def\ECCVSubNumber{5568}  %

\title{Supplementary}

\newenvironment{packed_enum}{
\begin{enumerate}
  \setlength{\itemsep}{1pt}
  \setlength{\parskip}{0pt}
  \setlength{\parsep}{0pt}
}{\end{enumerate}}

\titlerunning{Supplementary}
\author{Bram Wallace \and
Bharath Hariharan }
\authorrunning{B. Wallace and B. Hariharan}
\institute{Cornell University \\ \email{bw462@cornell.edu} } 

\maketitle

\setcounter{table}{0}
\renewcommand{\thetable}{S\arabic{table}}%
 \setcounter{figure}{0}
 \renewcommand{\thefigure}{S\arabic{figure}}%

\section{Experimental Setup}

\subsection{Architectures}

\subsubsection{ResNet26}

See Figure 3 and Section 3.2 of \cite{vdc1} for the original description.

\subsubsection{Autoencoder Generator}

The architecture is laid out in Table~\ref{table:generator}.

\begin{table}
\begin{center}
\begin{tabular}{ c  c}
\toprule
Layer & Output Shape  \\
\midrule
Input & 256 \\
   ConvTranspose2d-1    &        [-1, 512, 4, 4]        \\
       BatchNorm2d-2      &      [-1, 512, 4, 4]          \\
              ReLU-3     &       [-1, 512, 4, 4]             \\
   ConvTranspose2d-4     &       [-1, 256, 8, 8]    \\
       BatchNorm2d-5    &        [-1, 256, 8, 8]      \\
              ReLU-6     &       [-1, 256, 8, 8]               \\
   ConvTranspose2d-7    &      [-1, 128, 16, 16]     \\
       BatchNorm2d-8   &       [-1, 128, 16, 16]    \\
              ReLU-9  &        [-1, 128, 16, 16]    \\
  ConvTranspose2d-10      &     [-1, 64, 32, 32]      \\
      BatchNorm2d-11    &       [-1, 64, 32, 32]  \\
             ReLU-12     &      [-1, 64, 32, 32]      \\
  ConvTranspose2d-13      &      [-1, 3, 64, 64] \\
             Tanh-14    &        [-1, 3, 64, 64]     \\
\end{tabular}
\end{center}
   \caption{
Generator architecture. First convolution has stride of 1 and no padding, all subequent convolutions have stride of 2 with padding 1.
All kernels have size 4.
}
\label{table:generator}
\end{table}

\section{Training \& Evaluation}\label{sec:training}
All networks are trained using stochastic gradient descent for 120 epochs with an initial learning rate of 0.1 decayed by a factor of 10 at 80 and 100 epochs, with momentum of 0.9.
One addition to our training process was that of ``Earlier Stopping" for the Rotation and Jigsaw pretext tasks.
We found that even with traditional early stopping, validation accuracy could oscillate as the pretext overfit to the training data (especially in the Scenes \& Textures or Biological cases), potentially resulting in a poor model as the final result.
We stabilized this behavior by halting training when the training accuracy improves to 98\%, effect on accuracy shown in Table~\ref{table:early}.

\begin{table}
\centering
\caption{
Comparison of test accuracies with early stopping vs without.
Rotation in particular was stabilized and improved by this method.
Jigsaw was stabilized, but sometimes hampered.
For Jigsaw with less permutations than the 2000 reported the net effect was more positive.
The only qualitative difference in results was Jigsaw matching Instance Discrimination on the Internet domains instead of being outperformed. 
Both methods still fell far behind Rotation.
}
\begin{tabular}{ccccc}\toprule
& Jigsaw Early & Jigsaw Regular & Rotation Early & Rotation Regular \\
aircraft & 8 & 9 & 9 & 11 \\
cifar100 & 19 & 24 & 42 & 37 \\
cub & 9 & 9 & 12 & 14 \\
daimlerpedcls & 67 & 80 & 87 & 87 \\
dtd & 15 & 14 & 15 & 14 \\
gtsrb & 68 & 67 & 82 & 79 \\
isic & 57 & 59 & 60 & 62 \\
merced & 57 & 53 & 70 & 58 \\
omniglot & 18 & 24 & 46 & 54 \\
scenes & 33 & 33 & 42 & 40 \\
svhn & 50 & 53 & 80 & 78 \\
ucf101 & 25 & 22 & 42 & 45 \\
vgg-flowers & 22 & 19 & 23 & 22 \\
bach & 47 & 46 & 41 & 36 \\
protein atlas & 21 & 21 & 22 & 25 \\
kather & 79 & 78 & 57 & 61 \\
\bottomrule
\end{tabular}
\label{table:early}
\end{table}

\subsection{Dataset Splits}\label{sec:splits}

We use provided dataset splits when available, taking our validation data from training data when a train-validation split is not predetermined.\footnote{Despite using overlapping domains with the VDC, we are forced to use different splits in some cases due to the Visual Decathlon challenge not releasing the corresponding test labels.}
If no split was given, we generally used a 60-20-20 split within each class.
\textit{Full train-validation-test splits will be released along with our code and models}.

\subsection{Data Augmentation, Weight Decay, and Other Regularization}

A sensitive topic in any deep learning comparison is that of data augmentation or other forms of regularization, which can substantially alter performance.
In this work we are determined to give as fair of an apples-to-apples comparison as possible, and as such we apply minimal data augmentation and do not employ weight decay or other regularization methods.
The data augmentation used consists solely of resizing, random crops, and horizontal flips. 
Note that horizontal flips are not typically used on the symbolic domains, but are considered standard everywhere else.
We elected to go with the logical choice for 13 out of 16 of our domains, and employ horizontal flips in all of our main experiments.
We present results without flipping below.

\subsection{Effect of Horizontal Flipping on Symbolic}

As seen in Table~\ref{table:flips}, taking away horizontal flipping generally does not have major effects \textit{except} for improving Rotation-Omniglot substantially and hurting Jigsaw-SVHN significantly.
The former we attribute to the learning load of Rotation being used, while the latter we posit is due to the lack of horizontal flips allowing Jigsaw to use simpler cues for classification.

\begin{table}
\centering
\caption{
Each tuple is normal accuracy (with horizontal flips, as in paper) and accuracy without flips.
In general we see performance changes of only a few percentage points, qualitative comparisons largely hold.
The biggest differences are Rotation's improvement on Omniglot and Jigsaw's worsening on SVHN.
}
\begin{tabular}{cccccc}\toprule
& Autoencoding & Jigsaw & ID & Rotation & Supervised \\ \midrule
GTSRB &(57,58) & (66, 67) & (43, 39) & (82, 78) & (93, 93) \\
SVHN &(31, 33) & (55, 26) & (37, 34) & (80, 81) & (95, 95) \\
Omniglot & (18,19) & (26, 27) & (45, 47) & (46, 53)  & (79, 80) \\
\bottomrule
\end{tabular}
\label{table:flips}
\end{table}

\section{Implicit Dimensionality}

\begin{figure}
\centering
\includegraphics[width=\linewidth]{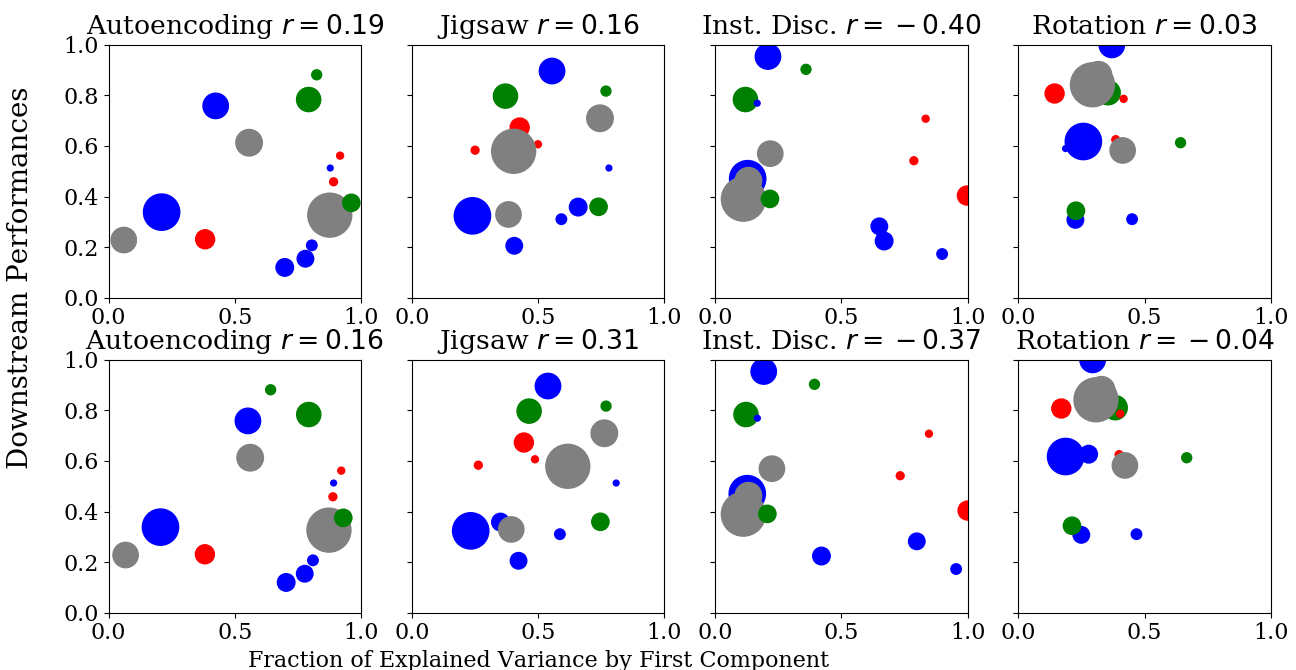}
   \caption{
Downstream normalized classification accuracy vs. the fraction of variance explained by the first component.
Top row is PCA on the entire training feature set, the bottom on validation.
The only moderately significant trends are those of Instance Discrimination, but we note that the trend holds with comparable strength for both sets.
}
\label{fig:pca_dim_vs_performance}
\end{figure}

We observe that the largest variations in explained variance between pretexts occur in the first dimension (Table~\ref{table:pca}), and investigate its use as a predictor in downstream performance.
Correlations are shown in Figure~\ref{fig:pca_dim_vs_performance}.
We do observe a moderate correlation between the explained variance in the first component and downstream normalized accuracy for Instance Discrimination.
While weak, this trend holds for PCA performed on both the training and validation images.
More significantly, we note the distinct separation formed around 0.5 on the x-axis and perform a t-test to determine that there is a moderately significant difference in downstream accuracies across this interval ($p=0.052$).
Thus implicit dimensionality is mildly predicitive of downstream performance for Instance Discrimination.

\begin{table}
\centering
\caption{
Fraction variance explained by the first $n$ values.
}
\begin{tabular}{ccccccccc}\toprule
& \multicolumn{2}{c}{Autoencoder} & \multicolumn{2}{c}{Jigsaw}& \multicolumn{2}{c}{Inst. Disc.} & \multicolumn{2}{c}{Rotation} \\
$n$ & 256 & 4096 & 256 & 4096 & 256 & 4096 & 256 & 4096  \\
\midrule
1 & 0.67 & 0.47 & 0.53 & 0.29 & 0.43 & 0.35 & 0.33 & 0.15 \\
2 & 0.75 & 0.54 & 0.69 & 0.39 & 0.51 & 0.41 & 0.51 & 0.23 \\
3 & 0.79 & 0.58 & 0.78 & 0.46 & 0.56 & 0.45 & 0.60 & 0.29 \\
4 & 0.82 & 0.61 & 0.83 & 0.50 & 0.61 & 0.48 & 0.67 & 0.33 \\
5 & 0.84 & 0.64 & 0.86 & 0.53 & 0.66 & 0.51 & 0.71 & 0.36 \\
10 & 0.89 & 0.72 & 0.92 & 0.63 & 0.79 & 0.61 & 0.82 & 0.46 \\
15 & 0.92 & 0.77 & 0.94 & 0.69 & 0.86 & 0.66 & 0.87 & 0.53 \\
20 & 0.94 & 0.81 & 0.95 & 0.72 & 0.90 & 0.70 & 0.90 & 0.57 \\
30 & 0.96 & 0.85 & 0.97 & 0.76 & 0.94 & 0.76 & 0.93 & 0.63 \\
40 & 0.97 & 0.88 & 0.97 & 0.79 & 0.96 & 0.79 & 0.95 & 0.67 \\
50 & 0.98 & 0.90 & 0.98 & 0.81 & 0.97 & 0.82 & 0.96 & 0.71 \\
60 & 0.99 & 0.92 & 0.98 & 0.83 & 0.98 & 0.84 & 0.96 & 0.73 \\
70 & 0.99 & 0.92 & 0.99 & 0.84 & 0.98 & 0.85 & 0.97 & 0.75 \\
80 & 0.99 & 0.93 & 0.99 & 0.85 & 0.99 & 0.86 & 0.97 & 0.77 \\
90 & 0.99 & 0.94 & 0.99 & 0.86 & 0.99 & 0.88 & 0.98 & 0.78 \\
100 & 0.99 & 0.94 & 0.99 & 0.87 & 0.99 & 0.88 & 0.98 & 0.80 \\
110 & 1.00 & 0.94 & 0.99 & 0.87 & 0.99 & 0.89 & 0.98 & 0.81 \\
120 & 1.00 & 0.95 & 0.99 & 0.88 & 0.99 & 0.90 & 0.99 & 0.82 \\
130 & 1.00 & 0.95 & 0.99 & 0.88 & 1.00 & 0.91 & 0.99 & 0.82 \\
140 & 1.00 & 0.95 & 1.00 & 0.89 & 1.00 & 0.91 & 0.99 & 0.83 \\
150 & 1.00 & 0.95 & 1.00 & 0.89 & 1.00 & 0.92 & 0.99 & 0.84 \\

\bottomrule
\end{tabular}
\label{table:pca}
\end{table}

\section{Nearest Neighbors}

Nearest neighbor examples are linked from the Github.

\section{Correlations of Pretexts with Downstream Accuracy}

\begin{figure}
\centering
\includegraphics[width=\linewidth]{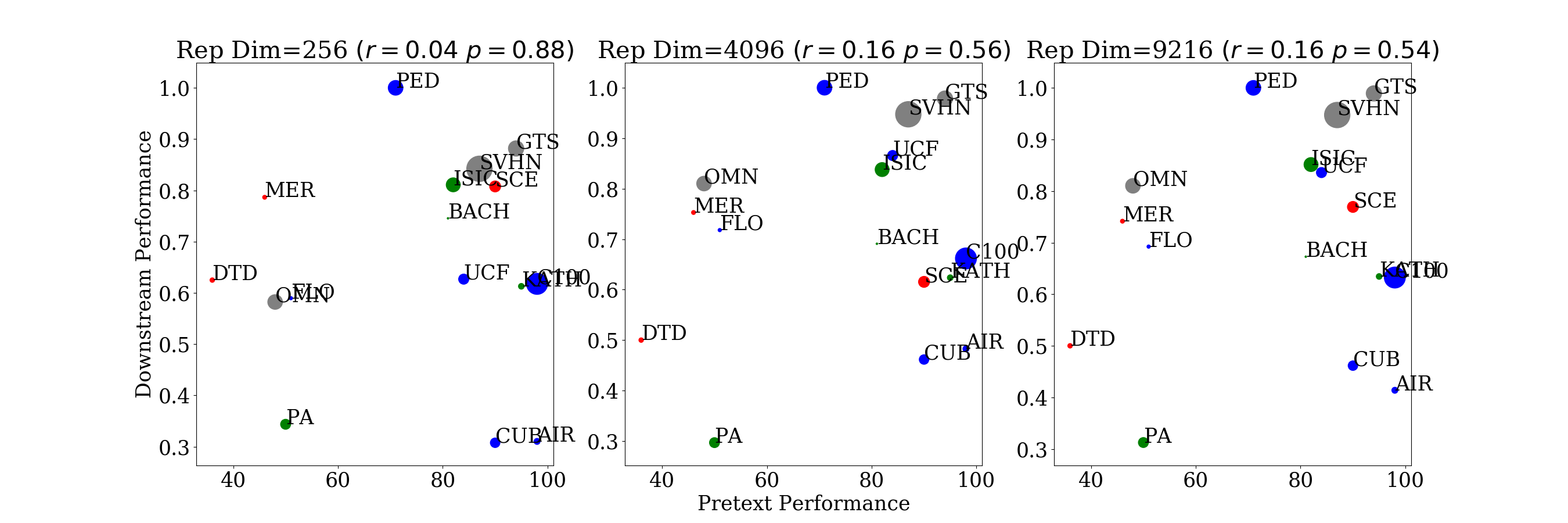}
   \caption{
Downstream normalized classification accuracy vs. performance on pretext task for Rotation.
}
\label{fig:rot}
\end{figure}

\begin{figure}
\centering
\includegraphics[width=\linewidth]{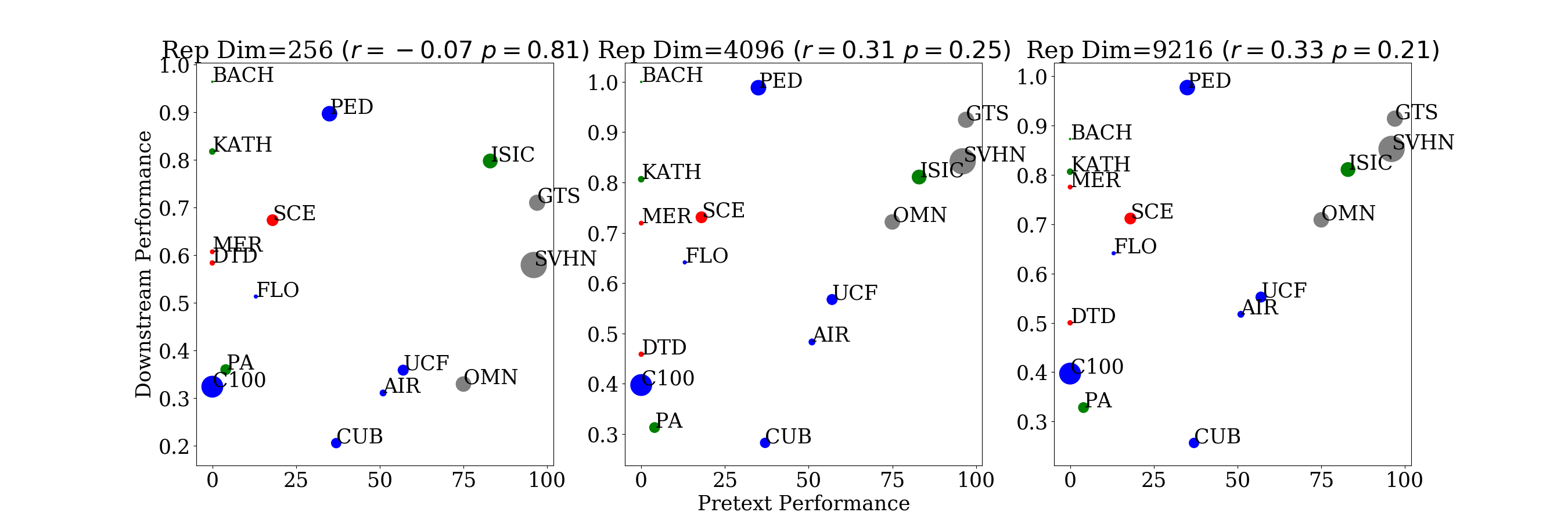}
   \caption{
Downstream normalized classification accuracy vs. performance on pretext task for Jigsaw.
}
\label{fig:jigsaw}
\end{figure}

\begin{figure}
\centering
\includegraphics[width=\linewidth]{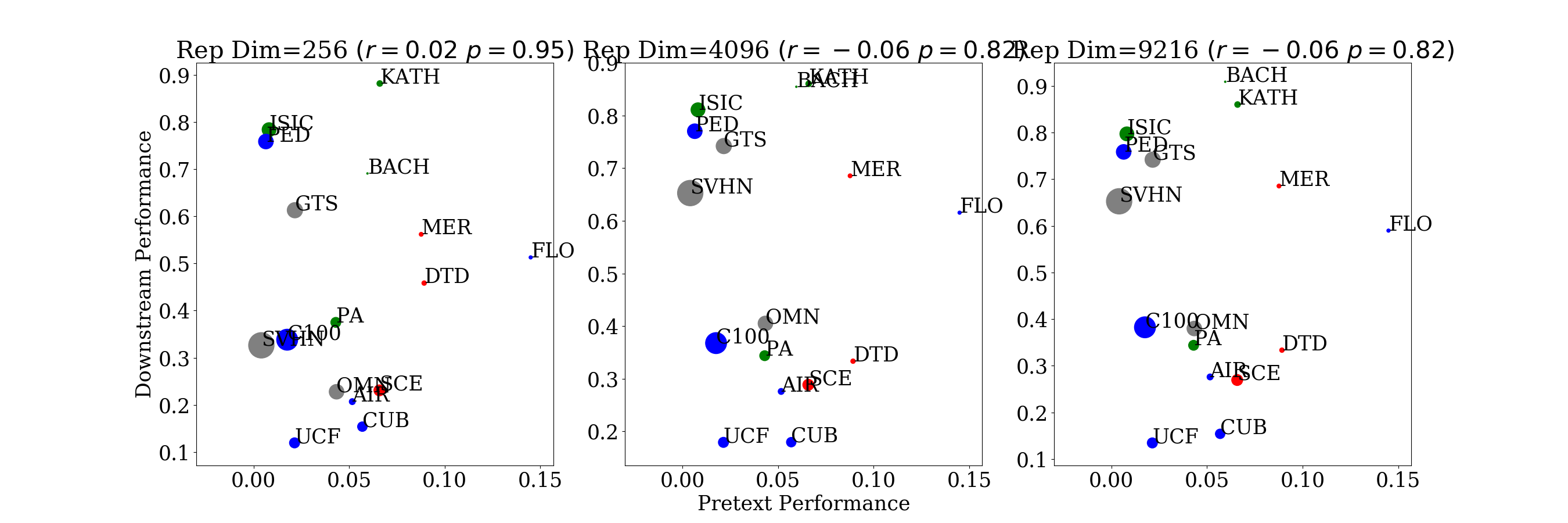}
   \caption{
Downstream normalized classification accuracy vs. performance on pretext task for Autoencoding.
}
\label{fig:ae}
\end{figure}

\begin{figure}
\centering
\includegraphics[width=\linewidth]{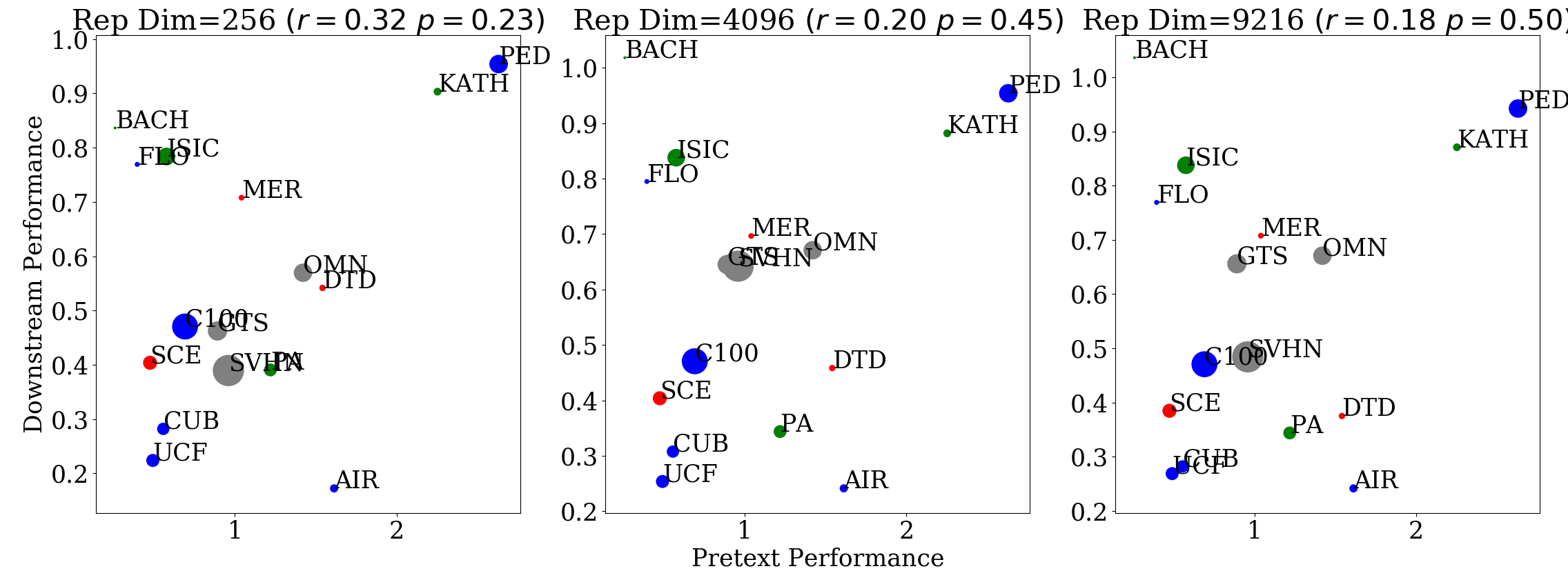}
   \caption{
Downstream normalized classification accuracy vs. performance on pretext task for Instance Discrimination.
}
\label{fig:inst_disc}
\end{figure}

Correlations for each task are shown in Figures~\ref{fig:rot},~\ref{fig:ae},~\ref{fig:jigsaw},~\ref{fig:inst_disc}.
X-axis is accuracy for Rotation/Jigsaw, loss of Autoencoding and ID.

{\small
\bibliographystyle{splncs04}
\bibliography{supp}
}